# MODELING AND DESIGN OF LONGITUDINAL AND LATERAL CONTROL SYSTEM WITH A FEED FORWARD CONTROLLER FOR A 4 WHEELED ROBOT

Younes El koudia, Jarou Tarik, Abdouni Jawad, Sofia El Idrissi and Elmahdi Nasri

Advanced Systems Engineering Laboratory, National School of Applied Sciences, Kenitra, Morocco

*ABSTRACT*

*The work show in this paper progresses through a sequence of physics-based increasing fidelity models that are used to design the robot controllers that respect the limits of the robot capabilities, develop a reference simple controller applicable to a large subset of tracking conditions, which include mostly non-invasive or highly dynamic movements and define path geometry following the control problem and develop both a simple geometric control and a dynamic model predictive control approach. In this paper, we propose for a nonlinear model with disturbance effect, the mathematical modeling of the longitudinal and lateral movements using PID with a feed-forward controller. This study proposes a feed-forward controller to eliminate the disturbance effect.*

*KEYWORDS*

*Robot, tracking, path geometry, geometric control, predictive control, feed-forward controller*

## 1. INTRODUCTION

For a robot to move around a track while staying within the lane, a PID controller modulization accomplished by calculating the velocity and the steering angle that is proportional to the lateral distance between the robot and the reference trajectory, which is the error of the cross track [1]. Effective speed and steering control across a range of speeds is necessary for the autonomous operation of mobile robots, although some of the reviewed, speed and steering automatic controllers in the literature have been implemented separately on production robots [2]. For an internal electric motor that is controlled by the throttle and steering, we present a design and implementation for a simplified adaptive cruise control (ACC) and lane keeping assistance (LKA) due to significant engine dynamics nonlinearities [3].

The lane keeping assistance system is a control system that helps a motorist keep their vehicle in a clearly marked highway lane while traveling safely. This system operates when a vehicle swerves from a lane, and the LKA automatically corrects the steering without further driver input to bring the vehicle back into the lane [4]. On the other hand, a device known as the cruise control system regulates and maintains the vehicle's speed at a predetermined point. The driver issues a signal of command. The cruise control system sends a control signal to the actuators that control the vehicle's throttle valve. This keeps the vehicle's speed constant and controls the fuel injection in the engine [5]. In this paper we are using these two concepts in a 4 wheeled robot, in a more streamlined model that combine the notions.

Industrial automation makes extensive use of independently driven mobile robots with four wheels [6]. In order to make the kinematic control study for this robot easier, we decided to treat the robot's kinematics like those of a bicycle for lateral movements, assuming that only the front wheel can be steered and that both the front and rear wheels are combined into unique wheels at the center of the front and rear axles [7]. For longitudinal movements, we treated it like a two wheeled robot, assuming that there is neither lateral nor rolling slip [8] [9].

Modern automobiles' automatic speed and steering control, also known as cruise control and lane keeping system, are typically advised for use at speeds greater than 13m/s and 16 m/s, it is therefore challenging to develop a controller for both systems that works well at speeds below 13m/s, most actuators have significant torque fluctuations at speeds below 13 m/s, a nonlinear phenomenon that causes significant variation in engine speed, crankshaft angular speed and mostly the steering wheel rotation angle [10].

The focal disturbance effect considered in this paper is the road conditions, and the two main challenges in designing an effective speed and steering controller are the lack of a complete mathematical model that brings together the two systems, and using one feedforward controller to reduce the nonlinear nature of the robot dynamics and kinematics in both longitudinal and lateral movements, especially for the targeted low speed range of 1-13m/s. Both of these reasons make the use of classical control strategies, such as PID Controller not easy [11]. The fundamental idea behind feed-forward control is to measure significant disturbance variables and correct them before they disrupt the process in order to improve performance. Our system is most affected by road incline and steering effect, especially when the robot starts at zero speed, when this disturbance will be taken into account.

## 2. KINEMATIC MODELING

In the kinematic study, only the velocities are taken into account. The motion of a differentially driven mobile robot in the simulation is characterized namely by no lateral or longitudinal slippage as the kinematic constraints.

Figure 1 represents the kinematic model for the robot used for this study.

### 2.1. Unicycle Representation

Note $\{x_f, y_f\}$ a fixed coordinate system and $\{x, y\}$ a mobile frame linked to the robot.

Let $q^f = [x^f, y^f, \theta^f]^T$ be a point of the coordinate system $\{x_f, y_f\}$ and $q = [x, y, \theta]^T$ a landmark point for $\{x, y\}$.

Points $q^f$ and $q$ are related by the orthogonal matrix R($\theta$).

$$q^f = R(\theta)\, q$$

With:
$$R(\theta) = \begin{pmatrix} cos(\theta) & -sin(\theta) & 0 \\ sin(\theta) & cos(\theta) & 0 \\ 0 & 0 & 1 \end{pmatrix}$$

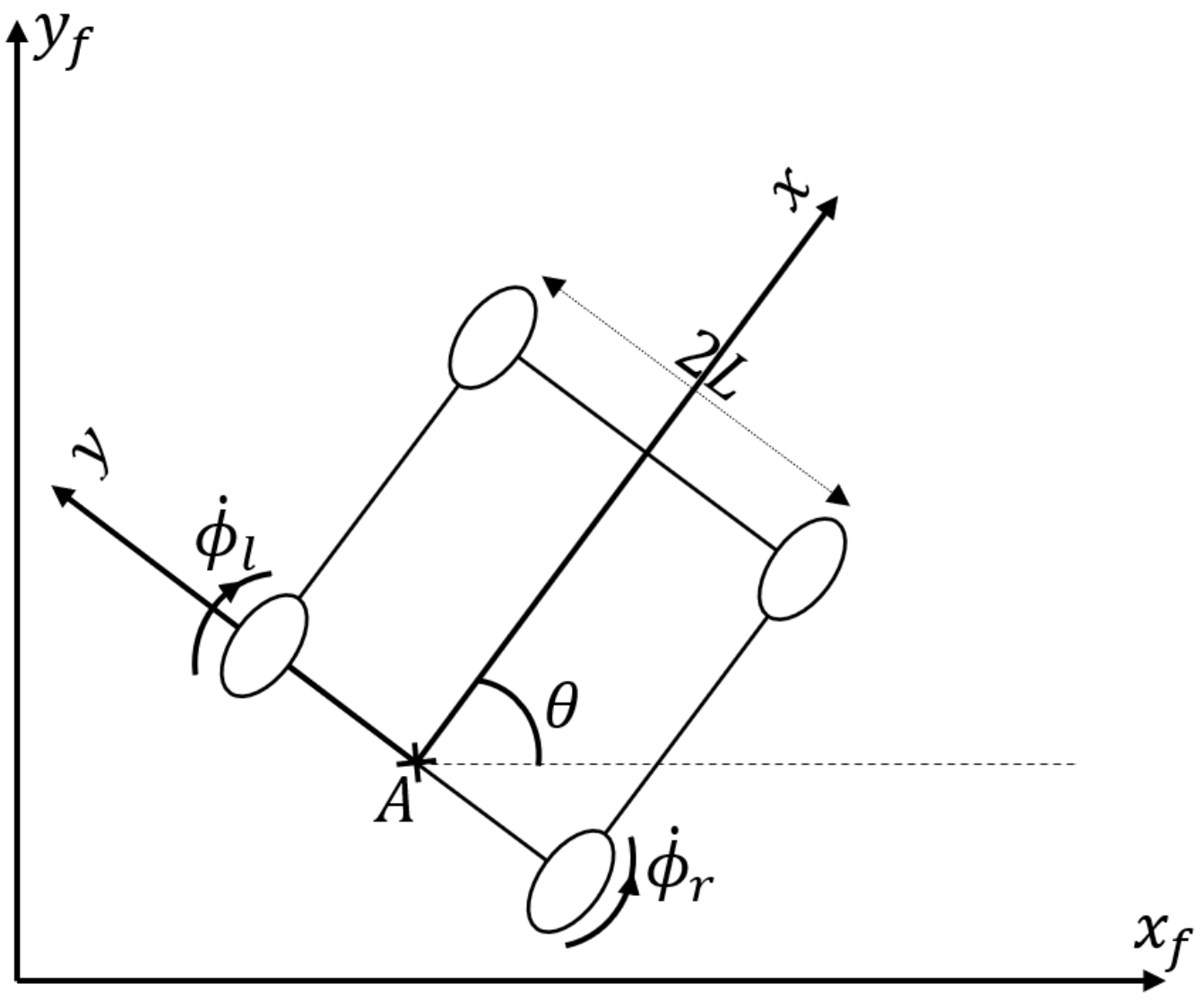

Figure 1 Kinematic Robot Model

- A : is the midpoint of the wheel axis.
- 2R : represents the diameter of the wheels.
- 2L : represents the robot width.
- $\dot{\phi}_r, \dot{\phi}_l$ : represent the rotational velocity of the right and left wheels, respectively.
- $\theta$ : is the angle of orientation of the robot.

### 2.1.1. Kinematic Constraints

The movement of the robot is characterized by two non-holonomic stresses that are obtained by two assumptions. A non-holonomic constraint is a non-integrable constraint involving the derivative with respect to time of the robot's coordinates [12]. If the robot can instantly move forward or backward but it cannot move right and left without the wheels slipping, it is said to have a non-holonomic constraint. On the other hand, if each wheel is able to move forward and sideways, it is said that this is a holonomic behavior of the robot.

### 2.1.2. Hypothesis

- Hypothesis 1: No Lateral Slip:

This constraint simply means that the robot can only move forward and backward, but not laterally. This means that the velocity of the robot associated with point A is zero along the lateral axis in the moving coordinate system, i.e. $\dot{y}_A = 0$

Using the rotation matrix $R(\theta)$, the expression of the robot velocity associated with point A in the fixed coordinate system is:

$$\begin{pmatrix} \dot{x}_A^f \\ \dot{y}_A^f \\ \dot{\theta}_A^f \end{pmatrix} = \begin{pmatrix} cos(\theta) & -sin(\theta) & 0 \\ sin(\theta) & cos(\theta) & 0 \\ 0 & 0 & 1 \end{pmatrix} \begin{pmatrix} \dot{x}_A \\ 0 \\ \dot{\theta}_A \end{pmatrix}$$

Then:

$$\begin{cases} \dot{x}_A^f = \dot{x}_A \cdot \cos(\theta) \\ \dot{y}_A^f = \dot{x}_A \cdot \sin(\theta) \end{cases}$$

Thus,

$$-\dot{x}_A^f \cdot \sin(\theta) + \dot{y}_A^f \cdot \cos(\theta) = 0$$

- Hypothesis 2: No Rolling Slip:

Non-slip rolling stress represents the fact that each wheel maintains a point in contact with the ground as it is shown in the Figure 2 below.

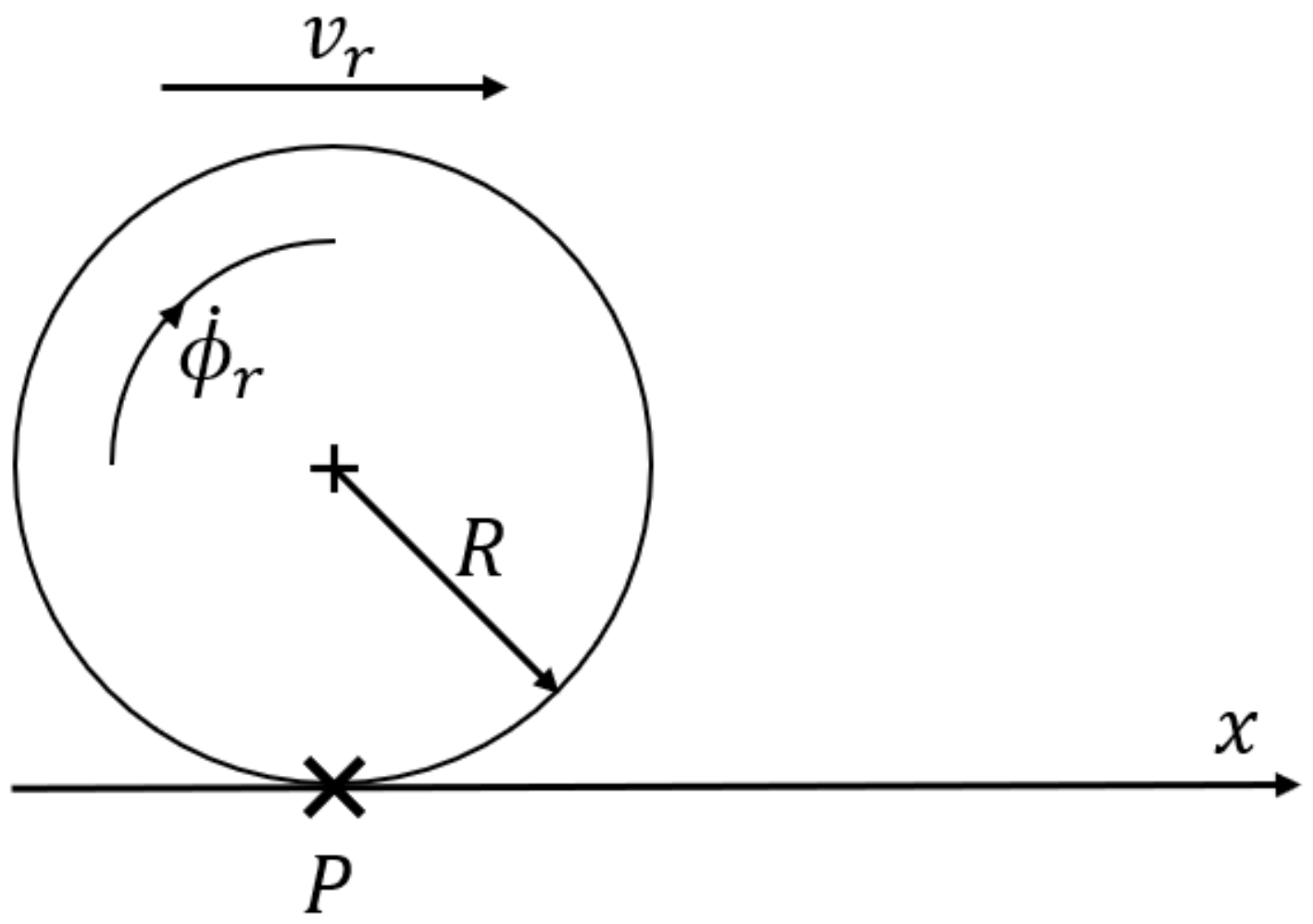

Figure 2 The right wheel

Thus, the linear velocity of each wheel of the robot at the point of contact P is given by:

$$v_{pr} = R\dot{\phi}_r$$

$$v_{pl} = R\dot{\phi}_l$$

Where $v_{pl}$ is the linear velocity of the left wheel, and $v_{pr}$ is the linear velocity of the right wheel.

The expressions of generalized positions and generalized velocities in the fixed coordinate system as a function of the coordinates of point A are given by:

Right wheel: $$\begin{cases} x_{pr}^f = x_A + L\sin(\theta) \\ y_{pr}^f = y_A - L\cos(\theta) \end{cases} \quad => \quad \begin{cases} \dot{x}_{pr}^f = \dot{x}_A + L\dot{\theta}\cos(\theta) \\ \dot{y}_{pr}^f = \dot{y}_A + L\dot{\theta}\sin(\theta) \end{cases}$$

Left wheel: $$\begin{cases} x_{pl}^f = x_A + L\sin(\theta) \\ y_{pl}^f = y_A - L\cos(\theta) \end{cases} \quad => \quad \begin{cases} \dot{x}_{pl}^f = \dot{x}_A - L\dot{\theta}\cos(\theta) \\ \dot{y}_{pl}^f = \dot{y}_A - L\dot{\theta}\sin(\theta) \end{cases}$$

Using the rotation matrix R($\theta$) and applying it to the right wheel we have:

$$\begin{pmatrix} \dot{x}_{pr}^f \\ \dot{y}_{pr}^f \\ \dot{\theta}^f \end{pmatrix} = \begin{pmatrix} cos(\theta) & -sin(\theta) & 0 \\ sin(\theta) & cos(\theta) & 0 \\ 0 & 0 & 1 \end{pmatrix} \begin{pmatrix} \dot{x}_{pr} \\ \dot{y}_{pr} \\ \dot{\theta} \end{pmatrix}$$

With $\dot{y}_{pr} = 0$ means that the velocity at point P of the right wheel is zero (because no lateral slip). Thus

$$\begin{pmatrix} \dot{x}_{pr}^f \\ \dot{y}_{pr}^f \\ \dot{\theta}^f \end{pmatrix} = \begin{pmatrix} \dot{x}_{pr} cos(\theta) \\ \dot{x}_{pr} sin(\theta) \\ \dot{\theta} \end{pmatrix}$$

We have: $$v_{pr} = \dot{x}_{pr} = R\dot{\phi}_r$$

Thus: $$\begin{cases} \dot{x}_{pr}^f \cos(\theta) = \dot{x}_{pr}\cos^2(\theta) & (a) \\ \dot{y}_{pr}^f sin(\theta) = \dot{x}_{pr}\sin^2(\theta) & (b) \end{cases}$$

By summing (a) and (b), we can form the equation system of two wheels:

$$\begin{cases} \dot{x}_{pr}^f \cos(\theta) + \dot{y}_{pr}^f sin(\theta) = R\dot{\phi}_r \\ \dot{x}_{pl}^f \cos(\theta) + \dot{y}_{pl}^f sin(\theta) = R\dot{\phi}_l \end{cases}$$

Hypothesis 1 and 2 and the previous equations produce the following constraints:

$$\begin{cases} -\dot{x}_A^f \cdot \sin(\theta) + \dot{y}_A^f \cdot \cos(\theta) = 0 \\ \dot{x}_{pr}^f \cos(\theta) + \dot{y}_{pr}^f sin(\theta) = R\dot{\phi}_r \\ \dot{x}_{pl}^f \cos(\theta) + \dot{y}_{pl}^f sin(\theta) = R\dot{\phi}_l \end{cases} \quad (1)$$

Then we can write:

$$A(q)\dot{q} = 0$$

$A(q)$ is the matrix of non-holonomic constraints given by:

$$A(q) = \begin{pmatrix} -\sin(\theta) & \cos(\theta) & 0 & 0 & 0 \\ \cos(\theta) & \sin(\theta) & L & -R & 0 \\ \cos(\theta) & \sin(\theta) & -L & 0 & -R \end{pmatrix}$$

$\dot{q}$ represents the derivative of the generalized coordinate $q$, given by $\dot{q} = [\dot{x}_A, \dot{y}_A, \dot{\theta}, \dot{\phi}_r, \dot{\phi}_l]^T$

Then we obtain that the expression of the linear velocities of the right and left wheels at the point of contact P is written in the following form:

$$\begin{cases} v_{pr} = v_A + L\dot{\theta} \\ v_{pl} = v_A - L\dot{\theta} \end{cases}$$

With $v_A$ the velocity of the point A, $v_{pr}$ is the velocity of the right wheel at point P and $v_{pl}$ is the velocity of the left wheel at point P.

Putting:

$$\begin{cases} v = v_A \\ \dot{\theta} = \omega \end{cases} \quad \text{Et} \quad \begin{cases} v_{pr} = v_r \\ v_{pl} = v_l \end{cases}$$

We obtain the expression of the linear velocity $v$ and the angular velocity $\omega$ of the mobile robot as a function of the rotational velocities of the left wheel $\dot{\phi}_l$ and the right wheel $\dot{\phi}_r$

$$v = \frac{v_r + v_l}{2} = R\frac{(\dot{\phi}_r + \dot{\phi}_l)}{2}$$

$$\omega = \frac{v_r - v_l}{2L} = R\frac{\dot{\phi}_r - \dot{\phi}_l}{2L}$$

In the moving coordinate system, the coordinates of point A are:

$$\begin{cases} \dot{x}_A^r = v \\ \dot{y}_A^r = 0 \\ \dot{\theta}_A^r = \omega \end{cases} \quad (2)$$

These equations represent the kinematic model of the unicycle robot.

### 2.2. Bicycle Model

The kinematic bicycle model reduces the left and right wheels to a pair of single wheels in the center of the front and rear axles, as shown in the Figure 3.

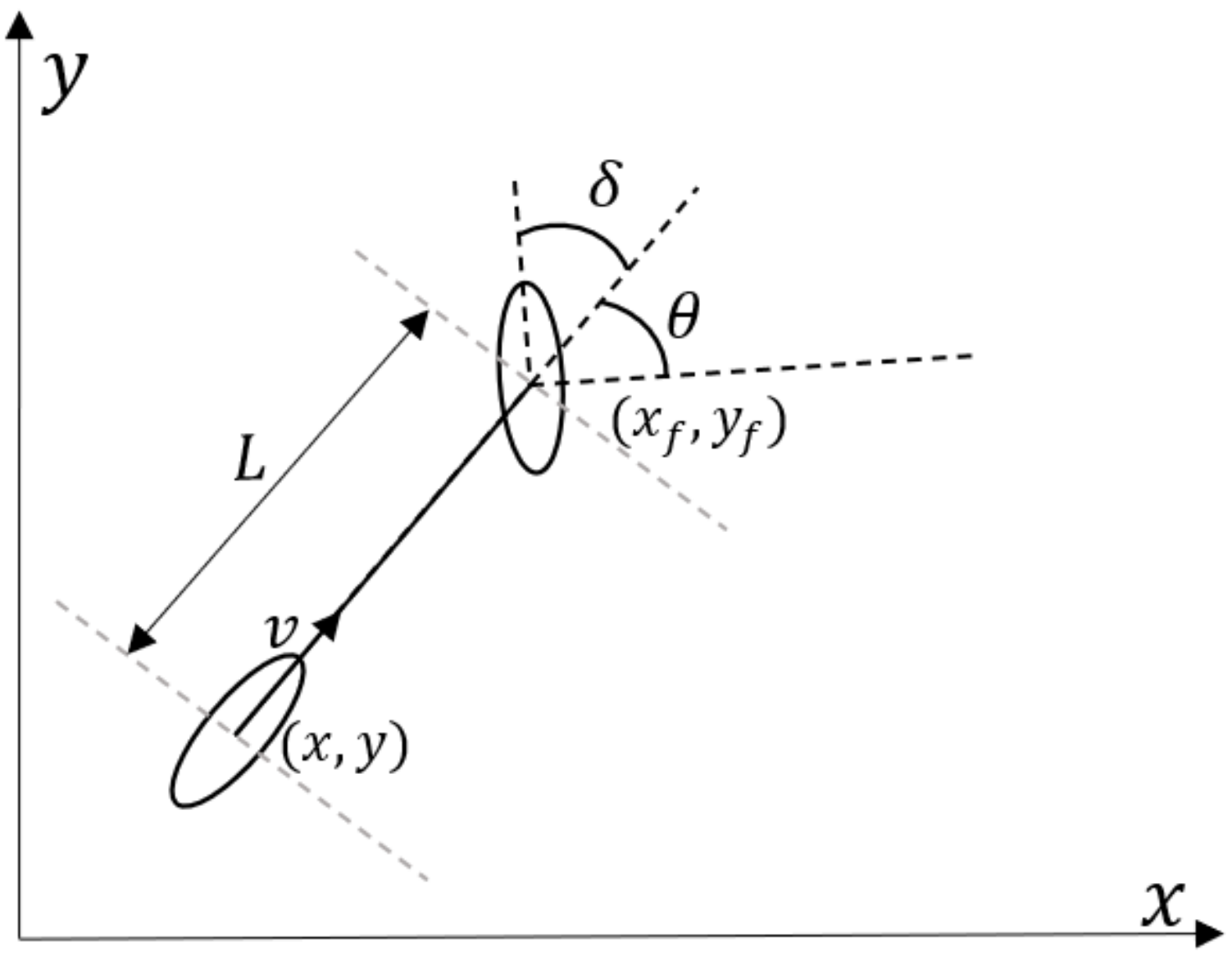

Figure 3 Bicycle model

The wheels are assumed to be side-free and only the front wheels are steerable [7]. By limiting the model to motion in a plane, the non-holonomic stress equations for the front and rear wheels are as follows:

$$\dot{x}_f \sin(\theta + \delta) - \dot{y}_f \cos(\theta + \delta) = 0$$

$$\dot{x} \sin(\theta) - \dot{y} \cos(\theta) = 0$$

Where $(x, y)$ is the overall coordinate of the rear wheel, $(x_f, y_f)$ is the overall coordinate of the front wheel, $\theta$ is the orientation of the robot in the overall framework, and $\delta$ is the steering angle in the body frame. As the front wheel is located at a distance L from the rear wheel according to the orientation of the trolley, $(x_f, y_f)$ can be expressed as follows,

$$\frac{d}{dt}\left(\dot{x}_f \sin(\theta + \delta) - \dot{y}_f \cos(\theta + \delta)\right) = 0$$

$$\frac{d(x + L\cos(\theta))}{\mathrm{d}t} \sin(\theta + \delta) - \frac{d(y + L\sin(\theta))}{\mathrm{d}t} \cos(\theta + \delta) = 0$$

$$\left(\dot{x} + L\dot{\theta}\cos(\theta)\right)\sin(\theta + \delta) - \left(\dot{y} + L\dot{\theta}\sin(\theta)\right)\cos(\theta + \delta) = 0$$

$$\dot{x}\sin(\theta + \delta) - \dot{y}\cos(\theta + \delta) - L\dot{\theta}(\sin^2(\theta)\cos(\delta) + \cos^2(\theta)\cos(\delta)) = 0$$

$$\dot{x}\sin(\theta + \delta) - \dot{y}\cos(\theta + \delta) - L\dot{\theta}(\cos(\delta)) = 0$$

By the elimination of $(x_f, y_f)$, the non-holonomy constraint on the rear wheel, is satisfied by $\dot{x}\cos(\theta)$ and $\dot{y}\sin(\theta)$ and any multiple scalars of these. This scalar corresponds to the longitudinal velocity $v$, such that,

$$\dot{x} = vcos(\theta)$$

$$\dot{y} = vsin(\theta)$$

Applying this to the stress on the front wheel gives a solution for $\dot{\theta}$,

$$\dot{\theta} = \frac{\dot{x}\sin(\theta + \delta) - \dot{y}\cos(\theta + \delta)}{L\cos(\delta)}$$

$$\dot{\theta} = \frac{v\cos(\theta)\,(\sin(\theta)\cos(\delta) + \cos(\theta)\sin(\delta))}{L\cos(\delta)} - \frac{v\sin(\theta)\,(\cos(\theta)\cos(\delta) - \sin(\theta)\sin(\delta))}{L\cos(\delta)}$$

$$\dot{\theta} = \frac{v(\cos^2(\theta) + \sin^2(\theta))\sin(\delta))}{L\cos(\delta)}$$

$$\dot{\theta} = \frac{v(\tan(\delta))}{L}$$

The instantaneous radius of curvature R of the robot determined from $v$ and leads to the previous introduction $\dot{\theta}$,

$$R = \frac{V}{\dot{\theta}}$$

$$\frac{v(\tan(\delta))}{L} = \frac{v}{R}$$

$$\tan(\delta) = \frac{L}{R}$$

$$\begin{bmatrix} \dot{x} \\ \dot{y} \\ \dot{\theta} \\ \dot{\delta} \end{bmatrix} = \begin{bmatrix} \cos(\theta) \\ \sin(\theta) \\ \frac{\tan(\delta)}{L} \\ 0 \end{bmatrix} v + \begin{bmatrix} 0 \\ 0 \\ 0 \\ 1 \end{bmatrix} \dot{\delta} \tag{3}$$

Where $v$ and $\dot{\delta}$ are respectively the longitudinal velocity and the angular velocity of the steering wheel.

### 2.3. Dynamic Modeling

In this first part any depreciation is neglected [9]. As a model with one degree of freedom we are only interested in the translational movement of the robot.

The potential energy of the robot:

$$E_p = E_{P_P} + E_{Pe} = mgx + \frac{1}{2}kx^2$$

The kinematic energy of the robot:

$$E_c = \frac{1}{2}m\dot{x}^2$$

Applies the Lagrange formalism, we find by applying the Laplace transform:

$$ms^2x + mg + kx = 0$$

Then the transfer equation is of the form:

$$G\lfloor s \rfloor = x(s) = \frac{{}^{mg}/_{k}}{\frac{m}{k}s^2 + 1} \tag{4}$$

## 3. CONTROL

The basic concept of feed-forward control is to measure important disturbance variables and take corrective action before they disrupt the process in order to improve the performance result.

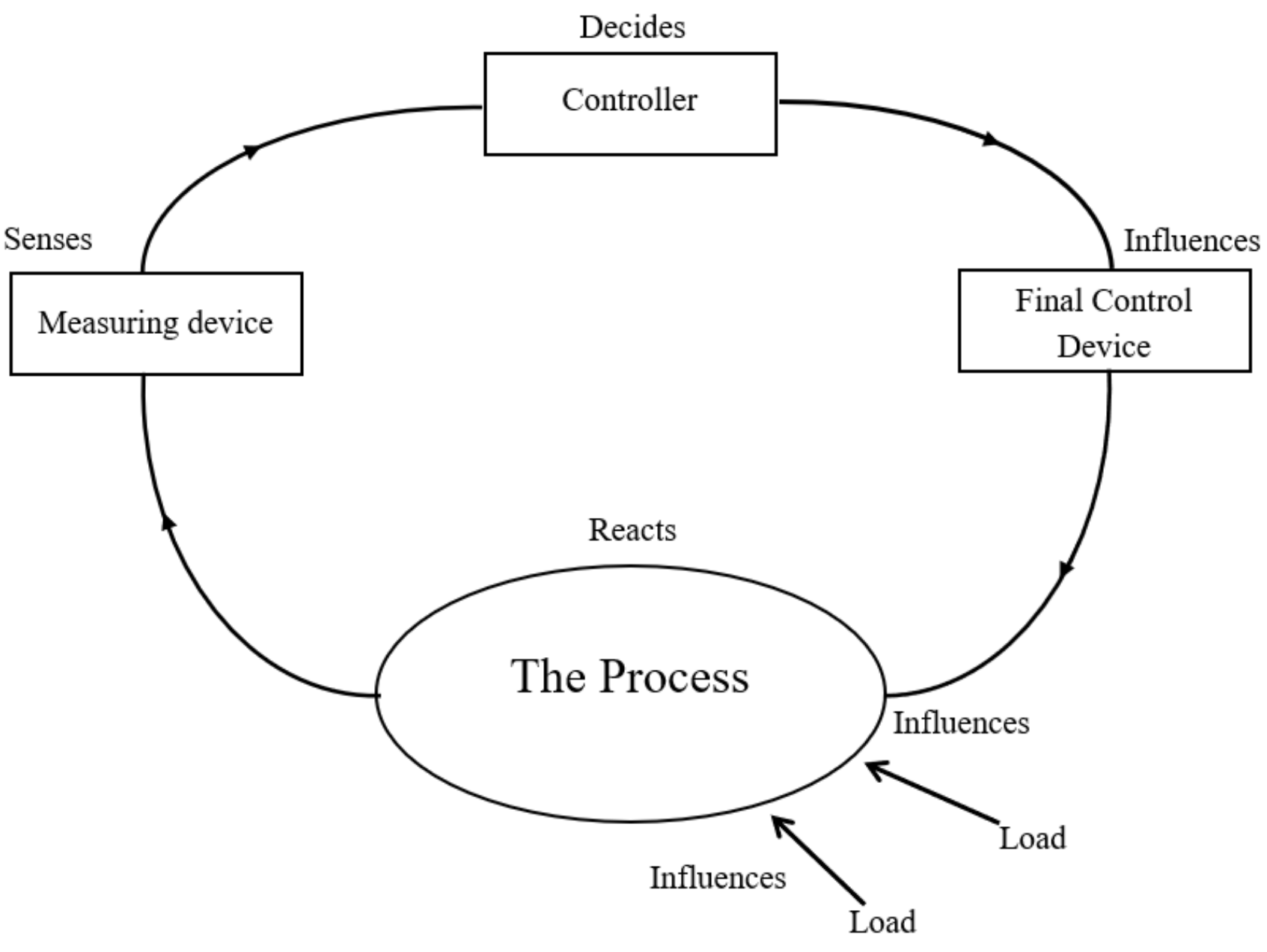

Figure 4 Advance control

If we break down the advance control into different stages, we obtain a thorough and comprehensive analysis of the planning and control system, a regular review of the system for input variables and interrelationships for a consolidated result, a collect data on input variables and synchronize them with the developed system, regularly analyze variations in actual input data compared to planned inputs and assess their effect on the expected result and a based on the analysis, take corrective actions to align planned and actual trajectories.

The main disturbance that acts on our system is the inclination of the road and the effect of the steering, especially when the robot starts at zero velocity, this disturbance is taken into account. Figure 5 shows an anticipatory control system, in which disturbances are measured and compensatory control actions are taken by the anticipatory controller.

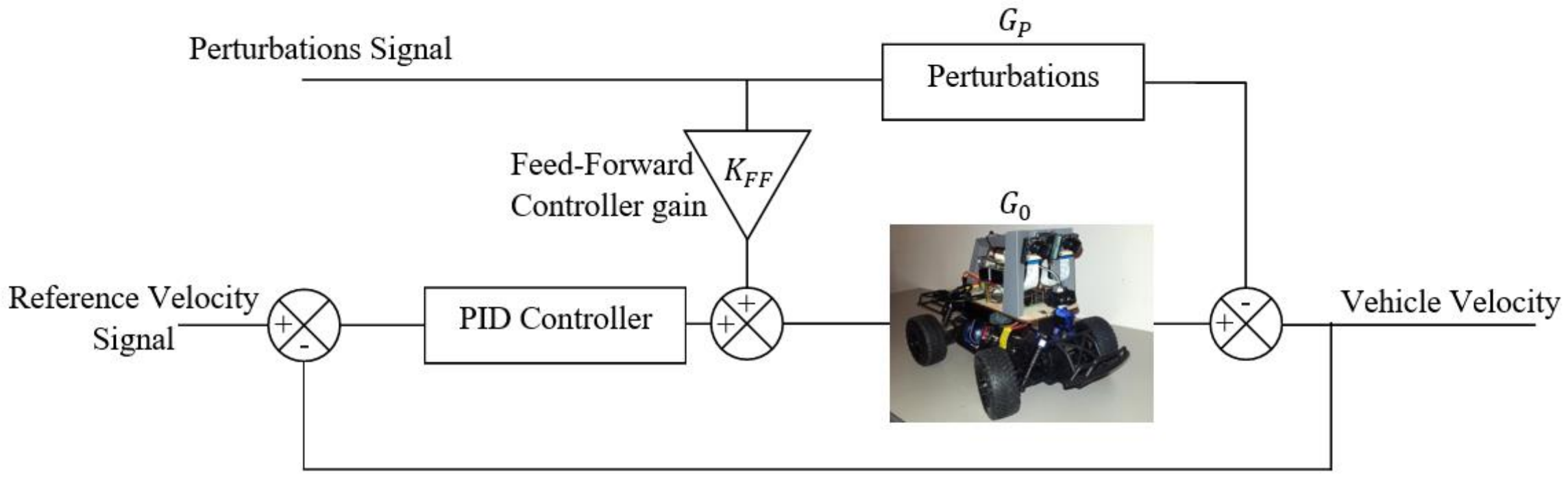

Figure 5 Block scheme for anticipation control

## 3.1. Representation of Dynamic Disturbances

The controller adjusts the motor torque gain to increase or decrease the motor drive force $F_d$ in response to the command signal, which is the reference velocity, and the speed sensor's feedback signal. The longitudinal elements of the vehicle as administered by Newton's low is

$$F_d = m\frac{\mathrm{d}v}{\mathrm{d}t} + F_a + F_g$$

$m\frac{\mathrm{d}v}{\mathrm{d}t}$ is the inertia force, $F_a$ is the aerodynamic resistance, $F_d$ the driving force of the engine and Fg is the resistance to rise (or the force of descent).

$$F_g = mg \sin\theta$$

$$F_a = c_a(v - v_w)^2$$

$v_w$ is the velocity of the wind thrust, $M$ is the mass of the carriage robot, $\theta$ is the slope of the road and $c_a$ is the coefficient of aerodynamic resistance.

Planning the direct control of this system starts with simplifying the model. Consideration is being given to setting all initial conditions to zero.

$$\dot{v} = \frac{1}{m}(F_d - C_a v^2)$$

$$TL(\ddot{v}) = TL\left(\frac{1}{m}\cdot\left(\dot{F}_d - 2c_a v\dot{v}\right)\right)$$

$$\mathrm{X}s^3 = \frac{1}{m}\cdot(sF_d - 2c_a \cdot X^2 s^3)$$

$$\mathrm{X}= \frac{c_a}{m}\left(-1 + \sqrt{1 - \frac{mF_d}{sc_a}}\right)$$

The basic concept of advance control (FF) is to measure important disturbance variables and take corrective action before they disrupt the process to improve performance. Especially when the robot starts at zero velocity, this disturbance will be taken into account.

A feed-forward control of the direct-acting control system is illustrated in the simulation part shown at the end of this article, where the disturbances are measured, and the compensatory command actions are taken by the direct-acting controller. Deviations in controlled variables can be calculated as follows:

$$\Delta v = G_p \cdot G_{FF} + G_0 = 0$$

$$G_{FF} = -G_p^{-1} \cdot G_0$$

With
$$G_P = \frac{c_a}{m}\left(-1 + \sqrt{1 - \frac{mF_d}{sc_a}}\right) \quad (5)$$

Et
$$G_0 = \frac{{mg}/{k}}{\frac{m}{k}s^2 + 1} \quad (6)$$

So, we find:

$$G_{FF} = \left(\frac{m}{c_a\left(1 - \sqrt{1 - \frac{mF_d}{sc_a}}\right)}\right)\cdot\left(\frac{{mg}/{k}}{\frac{m}{k}s^2 + 1}\right) \quad (7)$$

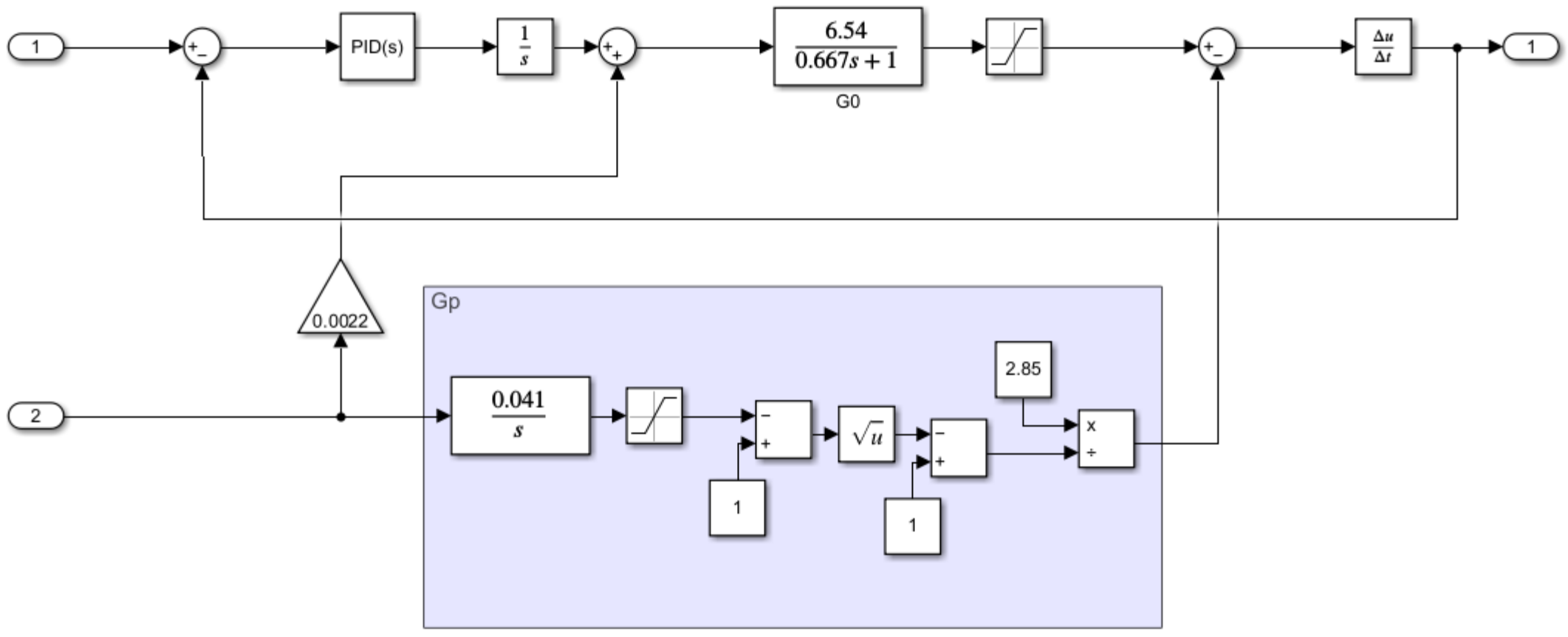

Figure 6 Dynamic robot model

### 3.2. Direct Kinematic Model (DKM)

The direct kinematic model allows us to know the Cartesian velocities $(v_x, v_y, v_z)$ of the tool and the rotational velocity vector $(\omega_x, \omega_y, \omega_z)$ of the tool coordinate system as a function of the positions and angular velocities of the axes.

From the kinematic modeling we found that:

$$v = \dot{x}_r = \frac{v_d + v_g}{2} = R\frac{\dot{\phi}_d + \dot{\phi}_g}{2}$$

$$\omega = \dot{\theta} = \frac{v_d - v_g}{2L} = R\frac{\dot{\phi}_d - \dot{\phi}_g}{2L}$$

From these equations we can build our direct kinematic model (DKM) under Simulink/Matlab

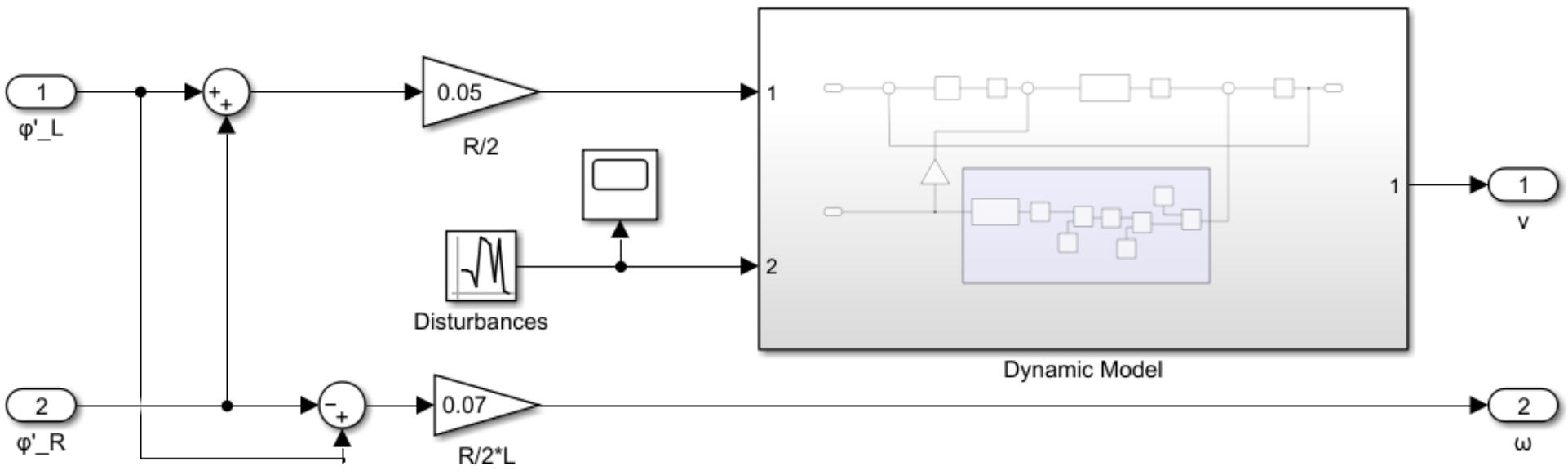

Figure 7  Direct kinematic model

### 3.3. Kinematic Model in the Robot Posture and the Lateral Control for Robot Orientation

We have modeled the equations $\dot{x}$, $\dot{y}$ et $\dot{\theta}$, note that $\gamma$ the angle that makes the frame with the x-axis, and $\theta$ the steering angle for the robot.

$$\dot{x} = v\cos(\theta),\ \dot{y} = v\sin(\theta),\ \dot{\theta} = \frac{v}{L}\tan(\gamma)$$

$$\dot{\gamma} = \omega_m \qquad (8)$$

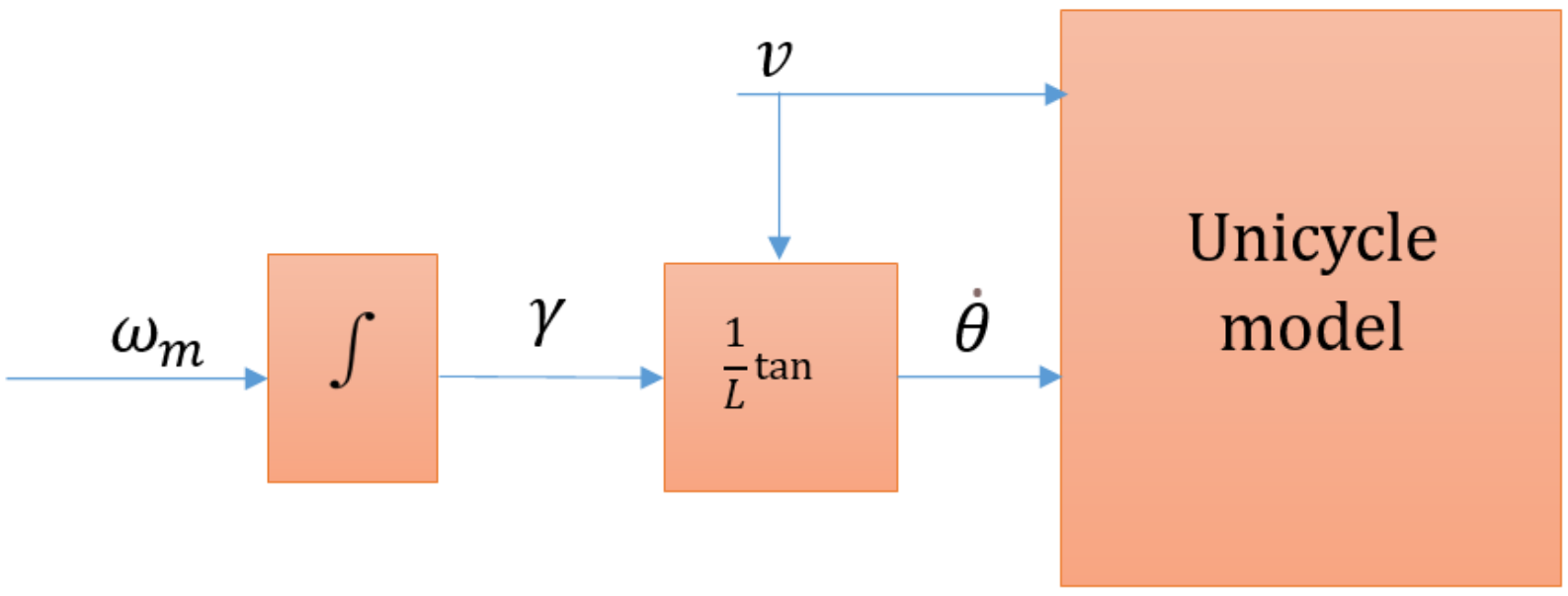

Figure 8  Block diagram corresponding to the unicycle model

In Simulink which allows us to have the velocities $\dot{x}$, $\dot{y}$ from $v$ and $\theta$.

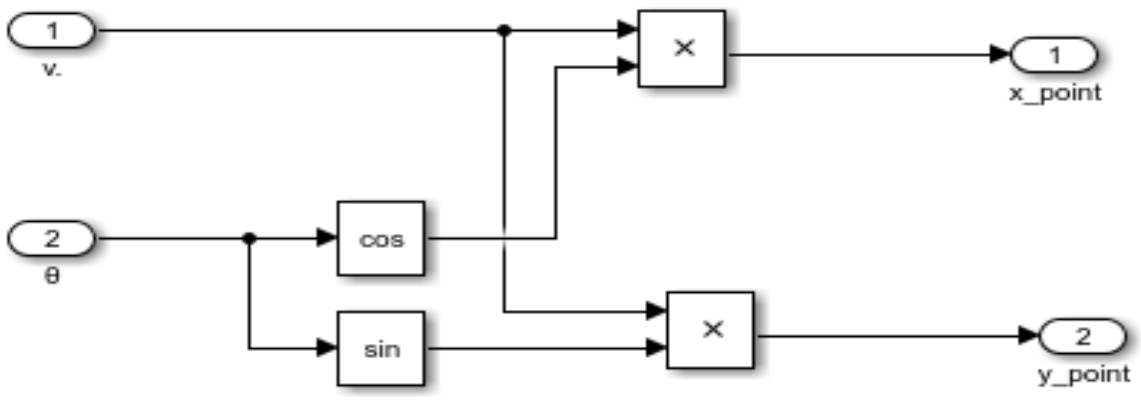

Figure 9 Kinematic model in robot posture

## 3.4. Inverse Kinematic Model (IKM)

The inverse kinematic model allows to switch from operational velocities $v$ and $\dot{\theta}$ to the velocities of each wheel.

The following equations are accepted:

$$\dot{\phi}_d = \frac{v + L\omega}{R} \tag{9}$$

$$\dot{\phi}_g = \frac{v - L\omega}{R} \tag{10}$$

From these relationships we can build our inverse kinematic model in Simulink.

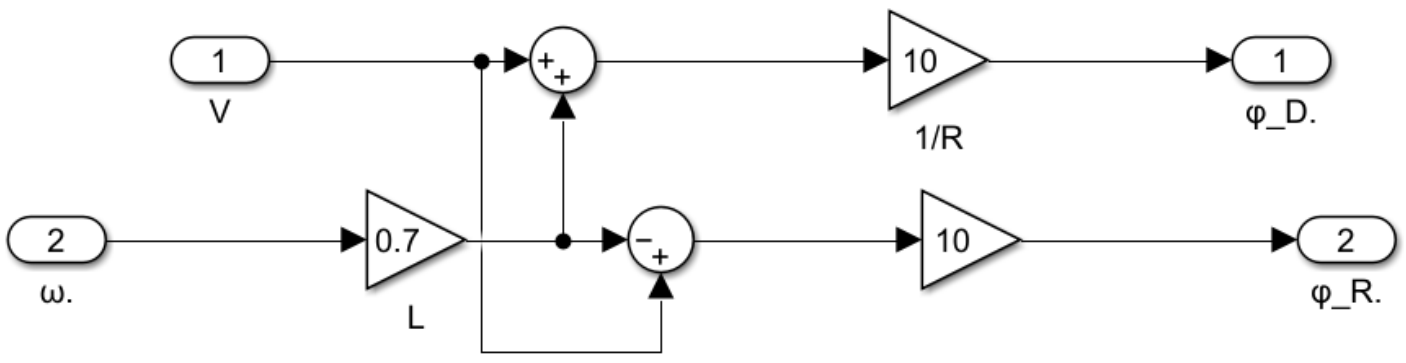

Figure 10 Inverse Kinematic Model

## 3.5. Actuator Modeling

DC motors are used to drive the wheels of our robot which are considered to be the actuators. The equations of a DC motor are the equation of:

The electrical circuit with the counter-electromotive force

$$L\frac{\mathrm{di}}{\mathrm{d}t} + r\mathrm{i} = u - v_b$$

The electromotive force

$$v_b = k_b\omega_m$$

The torque produced by the engine

$$\tau_m = k\mathrm{i}$$

Newton's second law for the motor shaft

$$J\ddot{\theta} + f\omega = \tau_m - \tau$$

The gear ratio

$$\omega_m = N\omega_R$$

The voltage $u$ is used as the input of the motor, $i$ is the armature current, the strength and inductance of the armature winding are respectively $(r, L)$. We consider $v_b$ the electromotive force, $\omega_m$ is the angular velocity of the robot and $\tau_m$ is the torque of the motor.

The torque constant and the electromotive force constant are respectively $(k, k_b)$. $J$ is the inertia of the engine and $f$ is the damping coefficient. $N$ is the reduction ratio

Table 1 shows the DC motor parameters and specifications that meet our robot design specifications

Table 1 DC motor parameters and specifications

| Parameter | Value |
|---|---|
| $r$ (Ω) | 0.8 |
| $L$ $(H)$ | 0.011 |
| $J\ (K_g.m^2)$ | 0.2 |
| $k\ (m.N/A)$ | 0.4 |
| $k_b\ (V.s/rad)$ | 0.4 |
| $N$ | 31.4 |

Below, in Figure 11, the Simulink model of the DC motor mathematical model.

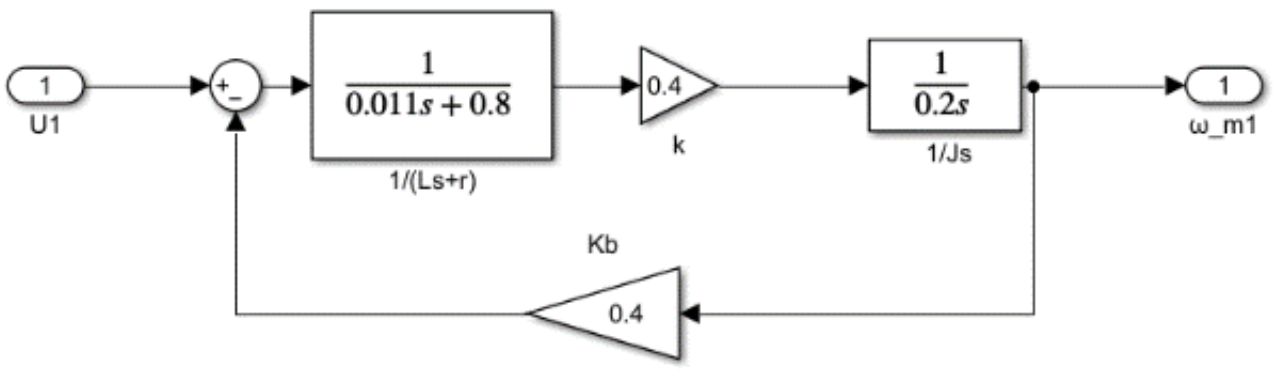

Figure 11  Modeling of motor actuators

With assembling all pieces and putting them in place, now we can determine the complete model of our robot.

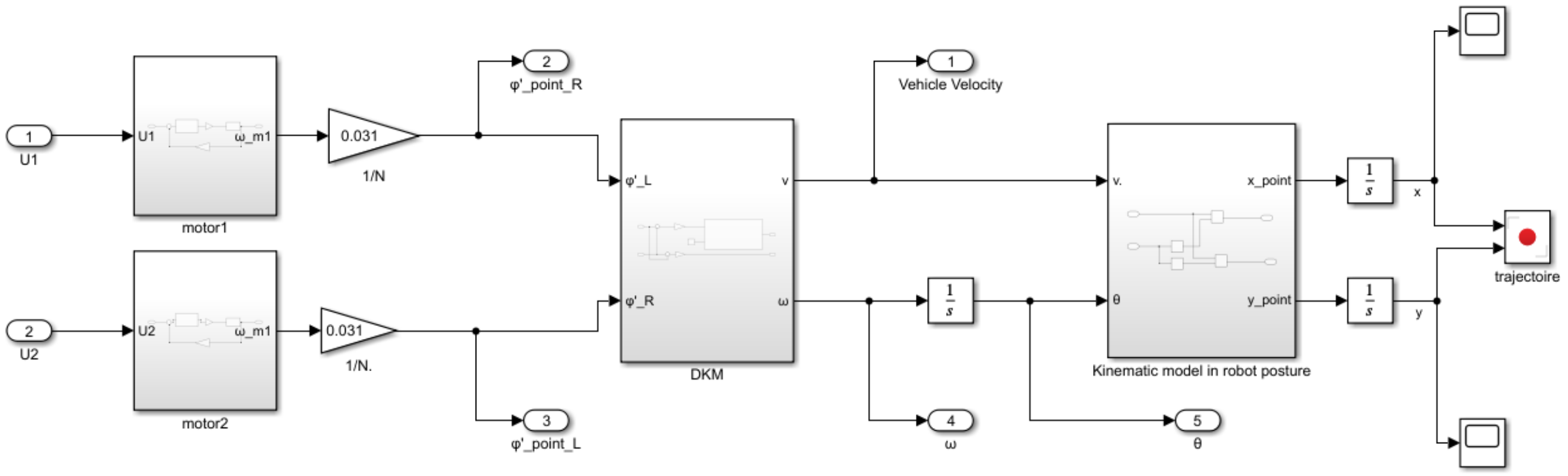

Figure 12 Unicycle Model

## 4. LONGITUDINAL AND LATERAL CONTROL OF THE ROBOT MOVEMENT

PID controller is a way to regulate an application to smoothly follow a value or a specific path. Although a full explanation of an application can be extensively complex, it summarized the math behind it in a super elegant and concise way and made the idea can easily extended to many real-world problems. In this chapter, we will follow the key structure of a PID controller.

### 4.1. Longitudinal Control

Longitudinal controller work in this case as a lane-keeping assist system, and it's designed to keep the vehicle in between two lanes. If the vehicle veers out of a lane, the controller must return the vehicle to its original position. Therefore, the goal of the PID controller is to reduce the amount of yaw error that occurs between the vehicle's heading and the centerline of the lane. The numerical model of PID regulator given by the mixed form of the PID transfer function written in the form shown here.

$$u(t) = k_p \mathrm{e}(t) + k_I \int_0^t \mathrm{e}(t)\,\mathrm{d}t + k_D \dot{\mathrm{e}}(t)$$

### 4.2. Lateral control

Pure tracking is the geometric trajectory tracking controller. A geometric path tracking controller is a controller that follows a reference trajectory using only the geometry of the carriage kinematics and the reference trajectory. The pure tracking controller uses an anticipation point which is a fixed distance on the reference trajectory forward of the carriage as follows. The carriage must move towards this point using a steering angle that we must calculate.

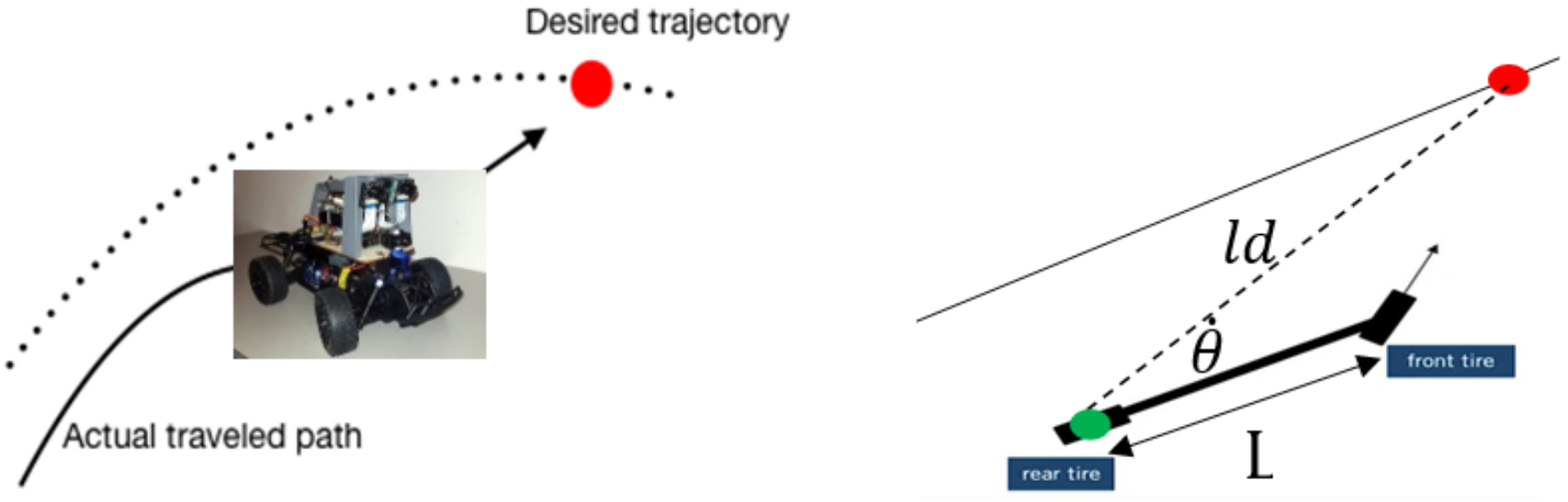

Figure 13 Pure Pursuit geometry

In this method, the center of the rear wheel is used as a reference point on the robot. The target point is selected as the red point in the figure above. And the distance between the rear axle and the target point is denoted $ld$. Our goal is to get the robot to a correct angle and then get to that point. The figure of geometric relationship is therefore as follows, the angle between the heading of the robot body and the line of sight is designated by $\dot{\theta}$. Because the robot is a rigid body and moves around the circle. The instant center of rotation *(ICR)* of this circle is shown as follows and the radius is denoted by R.

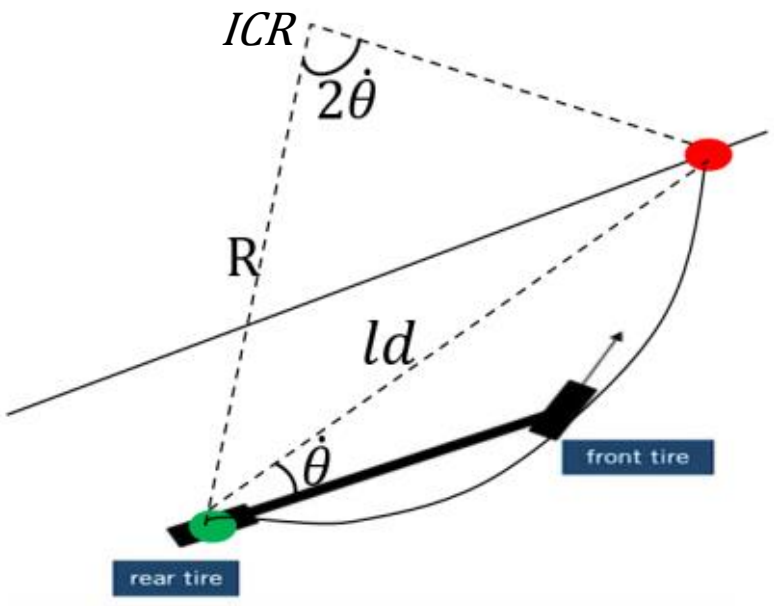

Figure 14 The angle between the carriage body heading and the line of sight

From the law of sines, we have

$$\frac{l_d}{\sin(2\dot{\theta})} = \frac{R}{\sin\left(\frac{\pi}{2} - \dot{\theta}\right)}$$

$$\frac{l_d}{2sin(\dot{\theta})\,cos(\dot{\theta})} = \frac{R}{cos(\dot{\theta})}$$

$$\frac{l_d}{sin(\dot{\theta})} = 2R$$

Thus

$$R = \frac{L}{tan(\delta)}$$

The steering angle $\delta$ can therefore be calculated as follows:

$$\delta = Arctan(\frac{2Lsin(\dot{\theta})}{l_d})$$

The pure tracking controller is a simple control. It ignores dynamic forces acting on the robot and assumes that the no-slip condition holds at the wheels. Also, if the controller is set for low speed, it would be dangerously aggressive at high velocities. An improvement is to vary the anticipation distance $ld$ according to the velocity of the robot.

To simplify the adjustment, the control law can be rewritten by scaling the anticipation distance according to the longitudinal velocity of the vehicle. Scaling the monitoring distance in this way is a common practice. In addition, the monitoring distance is usually saturated to a minimum and maximum value. In our case, these values are set at 1m and 3m respectively. It follows that the experiments are conducted on the lane change path.

$$\delta = Arctan(\frac{2Lsin(\dot{\theta})}{K_{ff} \cdot v}) \tag{11}$$

## 4.3. Setting of gains 'Ziegler Nichols Method'

The following is a breakdown of how each of the PID terms gains and govern the response of the regulator which is denoted by $u(t)$ and affects the steering angle of the robot:

- Proportional component:

When used by itself to calculate the steering angle, the proportional term produces a steering angle that is proportional to the Cross Track Error. On the other hand, it fluctuates around the reference trajectory as a result. The robot's rate of oscillation (or overshoot) around the reference trajectory is determined by the proportional coefficient ($k_P$).

- Derivative component:

The proportional component's overshoot is minimized by the derivative component by employing a rate of change of error. The robot's overshoot, or oscillation amplitude, distance from the reference trajectory can be optimized using this derivative coefficient ($k_D$) term.

- Integral component:

Systematic bias causes errors in the steering angle over time, which could eventually, but not immediately, cause the robot to leave the track. This issue is corrected by the integral component. The integral coefficient (Ki), which has a significant impact on the performance as a whole, should be carefully optimized in small steps because this component affects the error over time.

The Kp, Ki and Kd values could be manually adjusted through trial and error, but it would take a long time to get the robot to move smoothly around the track. Using an algorithm like twiddle to automatically tune the parameters is another option. While this might be the best strategy, it requires some time and effort to implement the algorithm. We chose a tuning method that was easier to use than the previous two: by comprehending the interdependencies of Kp, Ki, and Kd. We used the Ziegler-Nichols method and the equations in Table 2 below to select only two parameters to calculate the Kp, Ki, and Kd terms.

First, the integral action and the derivative action are canceled. The proportional action is increased until the output signal of the closed loop oscillates steadily. This gain is then noted $k_U$, This is the maximum gain (or critical gain). We note $T_u$ the period of oscillation of the signal. The parameters of the controller, $k_P$, $T_I$ and $T_D$ are chosen according to the following table.

Table 2 Ziegler Nichols' method

| Controller type | $k_P$ | $T_I$ | $T_D$ | $k_I$ | $k_D$ |
|---|---|---|---|---|---|
| P | $0.5k_U$ | | | | |
| PI | $0.45k_U$ | $T_U/1.2$ | | $0.54k_U/T_U$ | |
| PD | $0.8k_U$ | | $T_U/8$ | | $k_U T_U/10$ |
| PID | $0.6k_U$ | $T_U/2$ | $T_U/8$ | $1.2k_U/T_U$ | $3k_U T_U/40$ |

According to the modeling equations we have $k_U = \, {}^{mg}/_{k}$ and $T_U = T = \frac{2\pi}{\omega}$ while $\omega = \sqrt{\frac{k}{m}}$

$k_U$ = 6.54 and $T_U$ = 5.13s

So, we find that $k_P = 3{,}94$ ; $k_I = 1{,}52$ ; $k_D = 2{,}51$

## 5. SIMULATIONS AND INTERPRETATIONS

Many tests have done to examine the fidelity of this system, we are considering several initial values to show the simulated system using PID controller with feed-forward cancelling the effect of the disturbance from the road and figure 16 shows how is the system signal if we used the PID controller and if we did not use the PID controller it show that the signal is speed up under the reference speed but with PID controller it still in near the reference speed with small error. Figure 17 illustrates the real system signal during the driving on the flat and up and down hill with using PID controller without feed-forward and with feed-forward controller with 10m/sec reference speed. Table 3 shows the controller parameters and specifications that meet our design specifications.

Table 3 Controller Parameters

| Parameter | Value |
|---|---|
| $k_P$ | 3,94 |
| $k_I$ | 1,52 |
| $k_D$ | 2,51 |
| $k_{ff}$ | 0.0022 |

### 5.1. Absence and Presence of Controllers

Starting with the case where we do not have any type of controllers, giving 2.5m/s as the velocity reference generated by a pulse generator with 0.5rad as the steering angle, Figure 17 shows that the motors lose control, and the robot does not follow the reference. However, in the presence of the controllers we can notice that the velocity and intended angle serve as a reference for all cases and the results show us the significant rule of our controllers by demonstrating the stability of the system.

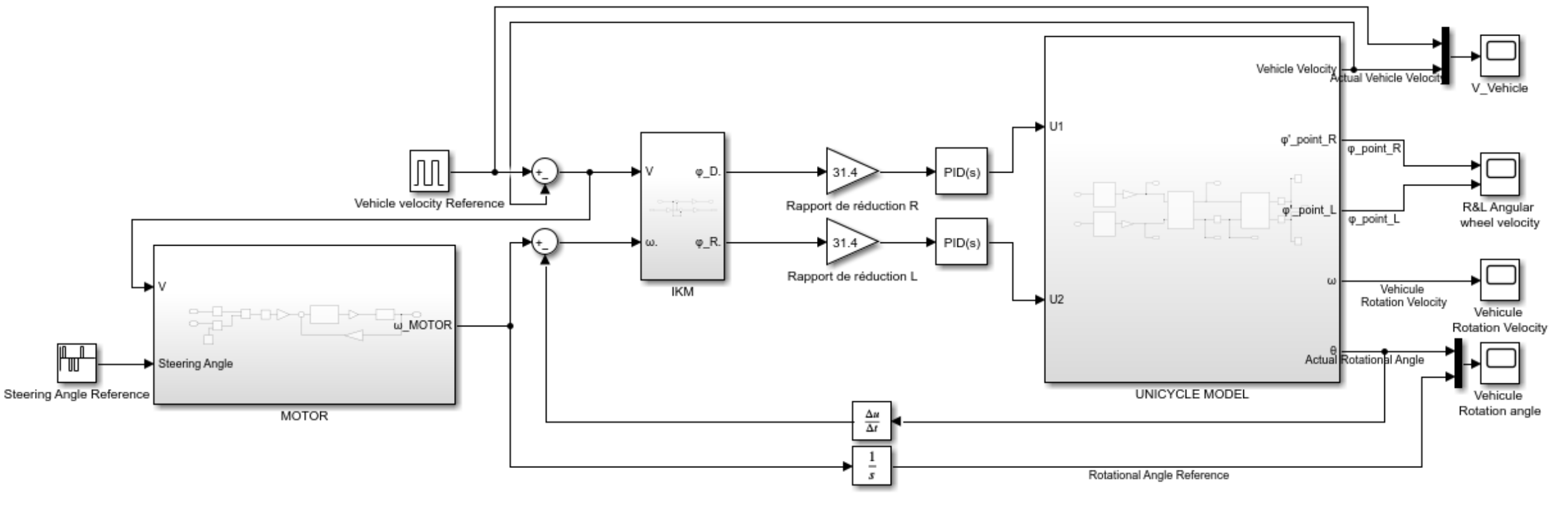

Figure 15 Presence of controllers

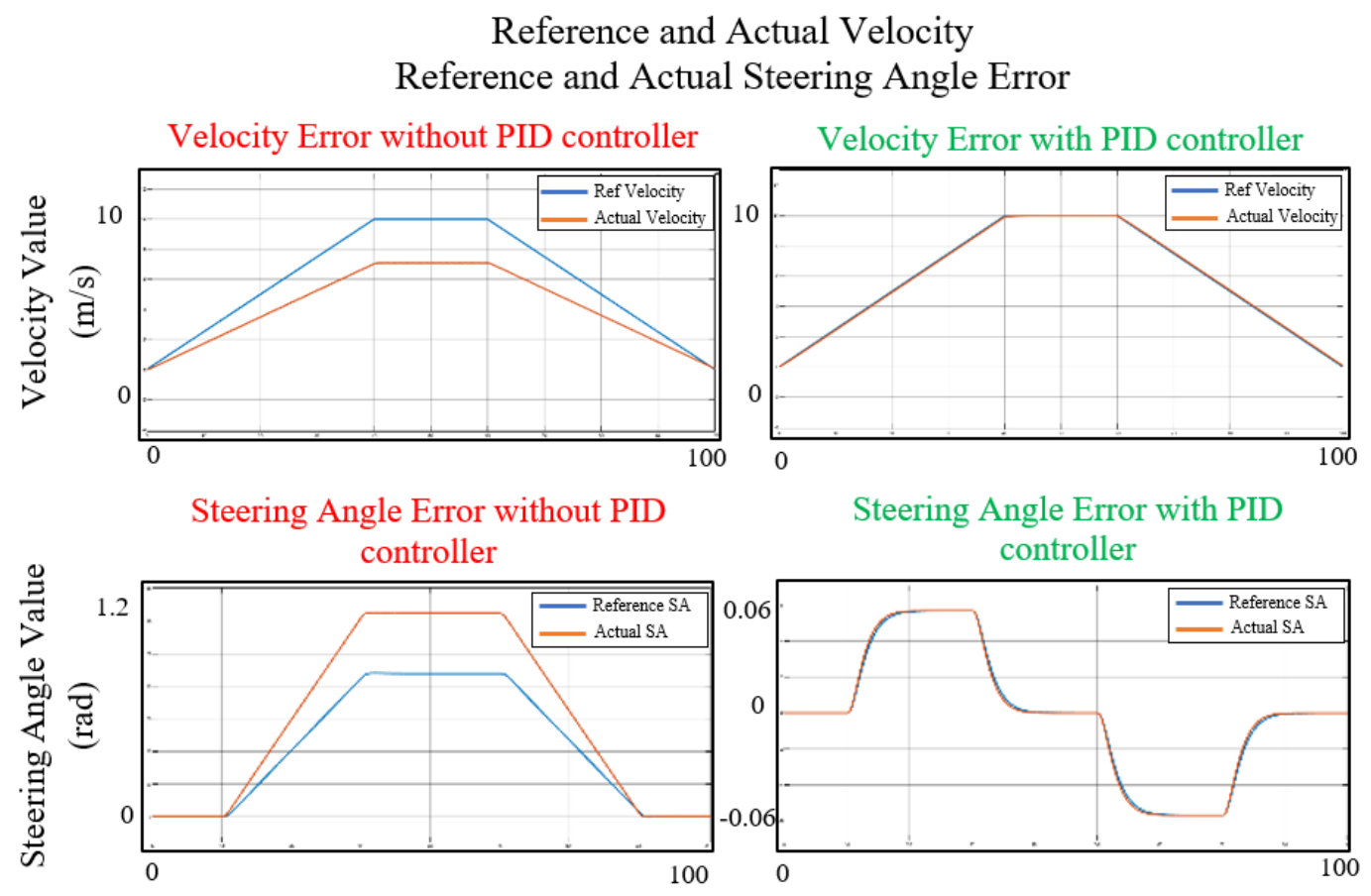

Figure 16 The real system with and without PID controller

## 5.2. Absence and presence of feedforward

The simulation in Figure 16 shows how the PID controller with feed-forward cancels the effect of the disturbance from the road incline and illustrates the real system signal and trajectory during the robot displacement, as well as the system without using any feedforward system to show how this disturbance can affect the trajectory of the robot. Many experiments have been conducted to obtain the best results for the comprehensive system of the two systems, ACC and LKA of the robot.

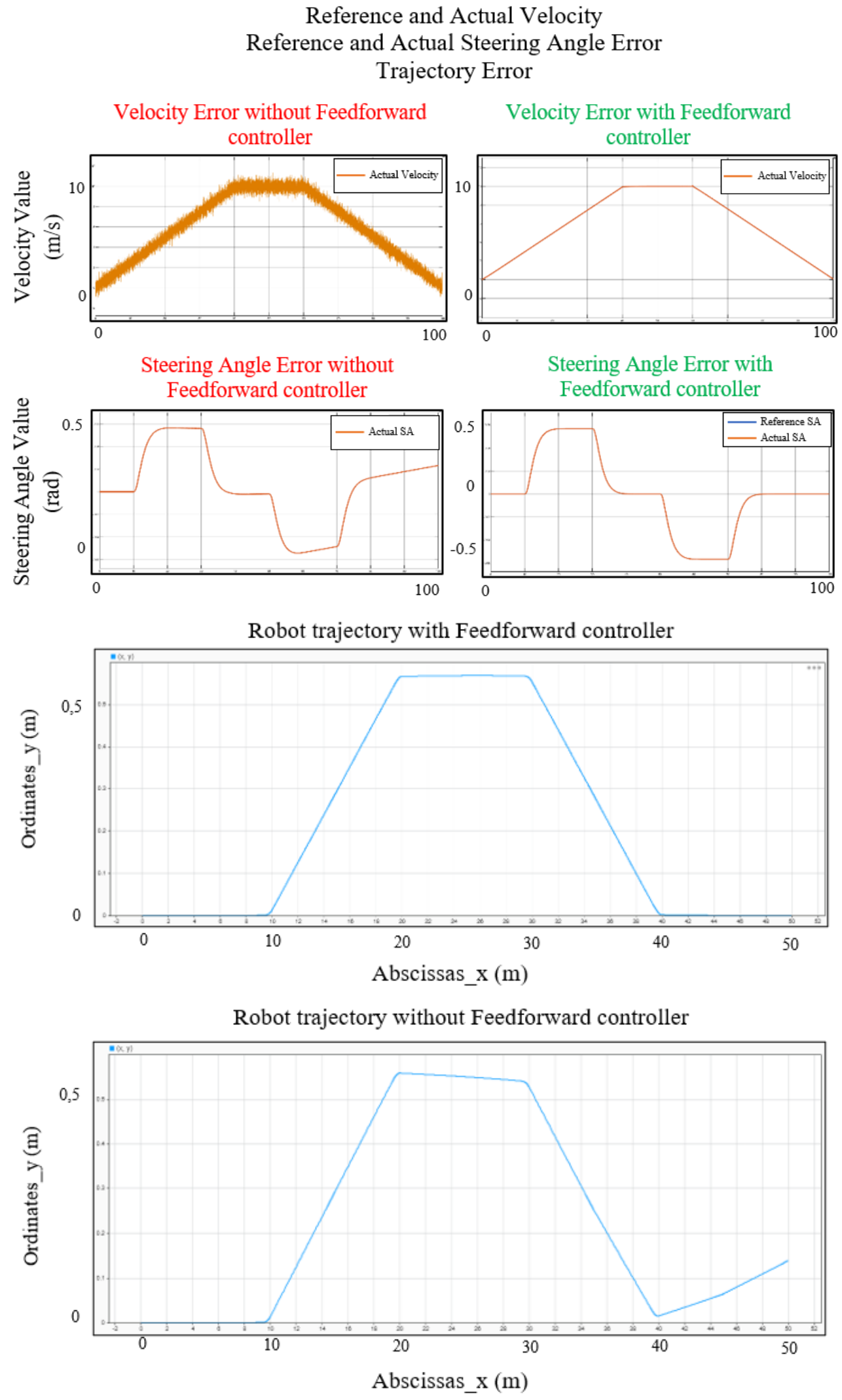

Figure 17 The real system with and without Feedforward controller

## 6. CONCLUSION AND FUTURE WORK

In conclusion, the mathematical unified model for the simplified ACC and LKA systems has been effectively derived inside this study by the implementation of a track algorithms for the ego-vehicle. The parameters for the PID and Feed-forward controllers that fulfill the system requirements are estimated, and the effect of disturbance from the incline road is simulated. However, in order to quantify the results, all of the responses were compared. Though, because the model outperformed previous control strategies, more advanced control systems will be used to test it in subsequent research.

## ACKNOWLEDGEMENTS

The authors would like to thank everyone, just everyone!

## AUTHOR

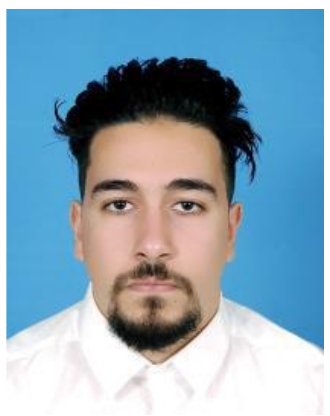

Younes EL KOUDIA a PhD student in the Electrical Department at the National School of Applied sciences in Kenitra, Morocco. Degree in Mechanical and automated systems Engineering from the National School of Applied sciences Fez, Morocco 2021. In the same year, he started his PhD within in the Engineering of the Advanced Systems Laboratory researching on Self-Driving Cars and can be contacted by email: younes.elkoudia@uit.ac.ma

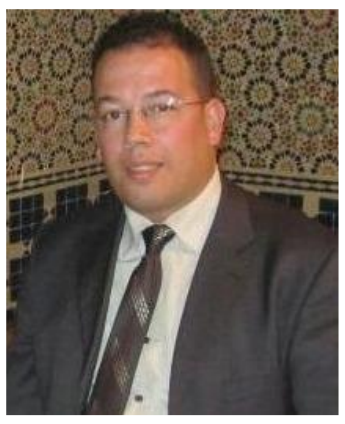

Tarik Jarou is University Professor at the National School of the Sciences Applied of the Ibn Tofail University, Kenitra, Morocco. He received Doctorat dregree (2008) in Electric Engineering from the Engineer School Mohammedia of the Mohamed V University , Rabat, Morocco. He is a member of the Advanced Systems Engineering Laboratory and Ex-member of LEECMS (Laboratory of Electric Engineering, Computing and Mathematical Sciences) of Ibn Tofail University, Kenitra, Morrocco. His main research area include the modelling, the control electronic and embedded systems for smart electric and cyberphysical system and their application fields in the automotive and aeronautical industry. He can be contacted at email: jaroutarik@hotmail.com.

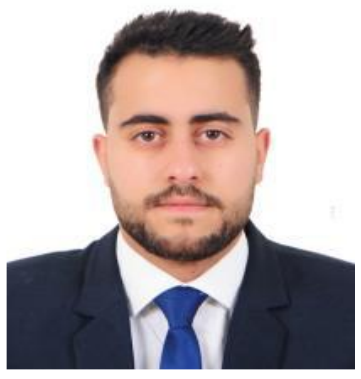

Jawad ABDOUNI is a PhD student at the Advance Systems Engineering Laboratory of the National School of Applied Sciences of Ibn Tofail University. He obtained his engineering degree in electromechanics from the Ecole Nationale Supérieure des Mines in Rabat in 2017. His field of research mainly focused on trajectory planning algorithms in autonomous navigation systems. He can be contacted at email: j.abdouni@enim.ac.ma.

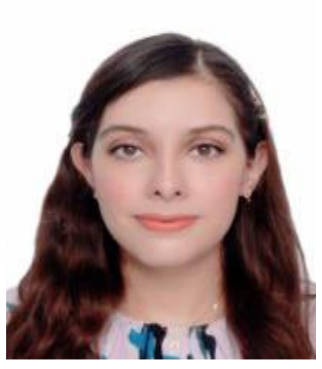

Sofia El Idrissi is a PhD student in the Advance Systems Engineering Laboratory at the National School of Applied Sciences within Ibn Tofail University. She received her engineering degree in embedded systems from the National School of Applied Sciences in 2019. Her field of research is mainly focused on autonomous vehicle’s motion planning and control systems in an urban environment. She can be contacted at email: sofia.elidrissi@uit.ac.ma

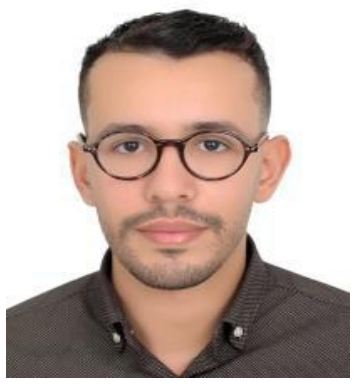

M. Nasri Elmehdi received his engineering degree in electical engineering from the national school of applied sciences Agadir Morocco, in 2019. He is now Ph.D. student in the Advanced systems Engineering Laboratory, National School of Applied Sciences,Ibn Tofail University /Kenitra, Morocco. His main research interest includes control of Lithium batteries, Battery Management system, power control in electrical vehicles. He can be contacted at email: elmahdi.nasri1@uit.ac.ma